\pdfoutput=1
\PassOptionsToPackage{xcdraw,table}{xcolor}

\documentclass[11pt]{article}

 \usepackage[final]{acl}

\usepackage{xspace}
\usepackage{times}
\usepackage{latexsym}
\usepackage{algorithm}
\usepackage{algpseudocode}
\usepackage{tikz}
\def\checkmark{\tikz\fill[scale=0.4](0,.35) -- (.25,0) -- (1,.7) -- (.25,.15) -- cycle;} 
\usepackage[utf8]{inputenc}
\usepackage{comment}

\usepackage{microtype}

\usepackage{inconsolata}
\usepackage{graphicx}
\usepackage{booktabs}

\newcommand{\spiritlm}{\textsc{Spiritlm}\xspace}

\newcommand{\llamatwobase}{\textsc{Llama-2}\xspace}
\newcommand{\llamathreebase}{\textsc{Llama-3}\xspace}
\newcommand{\llamatwochat}{\textsc{Llama-2-chat}\xspace}
\newcommand{\llamathreechat}{\textsc{Llama-3-chat}\xspace}
\newcommand{\llamatwo}{\textsc{Llama-2}\xspace}
\newcommand{\llamathree}{\textsc{Llama-3}\xspace}
\newcommand{\flores}{\textsc{FLORES-200}\xspace}
\newcommand{\mflores}{\textsc{2M-FLORES}\xspace}
\newcommand{\belebele}{\textsc{Belebele}\xspace}
\newcommand{\speechbelebele}{\textsc{2M-Belebele}\xspace}
\newcommand{\fleurs}{\textsc{FLEURS}\xspace}
\newcommand{\whisper}{\textsc{Whisper-large-v3}\xspace}
\newcommand{\whispers}{\textsc{Whisper}\xspace}
\newcommand{\mfourt}{\textsc{SeamlessM4T}\xspace}
\newcommand{\mms}{\textsc{MMS}\xspace}

\newcommand{\numlanguages}{\textsc{74}\xspace}

\title{\speechbelebele:
Highly Multilingual Speech and American Sign Language Comprehension Dataset
}

\author{Marta R. Costa-jussà, Bokai Yu, Pierre Andrews, Belen Alastruey, Necati Cihan Camgoz \\ \textbf{Joe Chuang, Jean Maillard, Christophe Ropers, Arina Turkantenko, Carleigh Wood} \\
FAIR, Meta\\
{\texttt{\{costajussa,bokai, mortimer,alastruey,neccam,}}\\{\texttt{joechuang,jeanmm,chrisropers,arinatur,carleighwood\}@meta.com}}}

\begin{document}
\maketitle
\begin{abstract}
We introduce the first highly multilingual speech and American Sign Language (ASL) comprehension dataset by extending \belebele. Our dataset covers \numlanguages spoken languages at the intersection of \belebele and \fleurs, and one sign language (ASL). 
We evaluate \speechbelebele dataset for both 5-shot and zero-shot settings and across languages, the speech comprehension accuracy is $\approx$ 2-3\% average lower compared to reading comprehension.
\end{abstract}

%We introduce the first highly multilingual speech and american sign language (ASL) comprehension dataset by extending to speech and ASL the \belebele reading comprehension dataset. Our dataset covers \numlanguages spoken languages at the intersection of \belebele and \fleurs plus ASL. We benchmark \speechbelebele dataset. For both 5-shot and zero-shot settings and across languages, the speech comprehension accuracy is {\color{red}20\%} average lower compared to reading comprehension accuracy.
%\end{abstract}

\section{Introduction}

%From the human perspective, speech comprehension is one of the relevant tasks for assessing communicative competence and it has been investigated in several contexts (e.g. in dialogues \cite{JONGMAN202195} or to qualify communication in classrooms \cite{PRODI2021108239}). Moreover, it has been shown that listening comprehension becomes the dominating influence on reading comprehension starting even in the elementary grades \cite{hogan:2014}.

From an AI perspective, text understanding and generation services are used globally in more than a hundred languages, but the scarcity of labeled data poses a significant challenge to developing functional systems in most languages. %It is expensive to extend them with human annotations.
Although natural language processing (NLP) datasets with extensive language coverage, such as \flores \citep{nllbteam2022language}, are available, they mainly concentrate on machine translation (MT). Multilingual evaluation benchmarks such as those for multilingual question answering \citep{lewis-etal-2020-mlqa,clark-etal-2020-tydi}, natural language inference \citep{conneau-etal-2018-xnli}, summarization \citep{hasan-etal-2021-xl,ladhak-etal-2020-wikilingua}, %(Ladhak et al., 2020;),
and reasoning datasets \citep{ponti-etal-2020-xcopa, lin-etal-2021-common} collectively cover only about 30 languages. Furthermore, the extension of such datasets to speech or American Sign Language (ASL) is lacking, with the exception of \fleurs \citep{conneau2022fleurs,tanzer2024fleursaslincludingamericansign}, which is based on \flores. 

The recent \belebele benchmark is the first corpus that addresses text reading comprehension for a large number of languages following a multi-way parallel approach \citep{bandarkar2023belebele}. The entire \belebele text statistics are summarized in Table \ref{tab:statistics}. Currently, there are no highly multilingual evaluation datasets for natural language understanding that cover either both speech and text, and/or ASL.

The outstanding performance of some MT and text-to-speech (TTS) models has enabled a rise in the number of works using synthetically generated training data. Furthermore, some recent works propose to also use synthetic data for evaluation; e.g., \cite{ustun2024aya, communication2023seamless, nguyen2024spiritlm, nachmani2023spoken}. This strategy allows researchers to extend datasets to low-resource languages and to other modalities, such as speech. However, we prove that using synthetic data for evaluation does not provide comparable conclusions as relying on human speech for the particular task of automatic speech recognition (ASR) and the \fleurs domain (see Appendix \ref{sec:ablation}).  

The evaluation dataset that is closest to the speech comprehension evaluation dataset presented in this paper is the generative QA dataset proposed in \cite{nachmani2023spoken}. The questions are taken from two sources: the WebQuestions dataset created by \citet{berant-etal-2013-semantic} and a new test set called ``LLama Questions.'' The dataset covers 300 questions in English. In this work, we extend the \belebele dataset to speech and sign (Section \ref{sec:dataset}). %Additionally, we report a partial extension to document-level. 
By doing so, we create the first highly multilingual speech and sign comprehension dataset\footnote{Appendix \ref{sec:appendix} specifies language coverage.}: \speechbelebele. Compared to spoken languages, sign languages are considered low-resource languages for natural language processing \citep{yin2021including}. Most popular datasets cover small domains of discourse; e.g., weather broadcasts \citep{Camgoz_2018_CVPR}, which has limited real world applications. There have been previous releases of large scale open domain sign language datasets; e.g., \cite{albanie2021bbc, shi2022open, uthus2024youtube}. However, the results and challenges on such datasets suggest that computational sign language research still requires additional datasets to reach the performance of their spoken language counterparts \citep{muller2022findings, muller2023findings}. With the release of the ASL extension of the \belebele dataset, we aim to provide additional, high quality sign language data with gloss annotations to underpin further computational sign language research. Furthermore, due to the paragraph-level nature of the \belebele dataset, we enable paragraph-context sign language translation, which has been reported to improve translation performance \citep{sincan2023context}. \speechbelebele is composed of human speech recordings covering \numlanguages languages and human sign recordings for ASL. 

As a by-product of \speechbelebele, we also extend the \fleurs dataset (which is widely used to benchmark language identification and ASR) by providing recordings for more \flores sentences than were previously available and adding sign language, creating a new \mflores. This \mflores extends \fleurs by 20\%.

Finally, we provide a very basic set of experiments that evaluate \speechbelebele and provide some reference results on the dataset. We use direct and/or cascaded systems to evaluate \speechbelebele dataset with direct and/or cascaded systems (Section \ref{sec:experiments}). We also list several further experimentation that \speechbelebele unblocks. Note that the main contribution of this paper is the creation of the first highly multilingual speech and sign comprehension dataset. The complete set of experiments is out of the scope of this paper (see Section on Limitations). By open-sourcing our dataset, we encourage the scientific community to pursue such experimentation.  

%We do not benchmark the ASL part of \speechbelebele because of the lack of an open-source model capable of doing so. 
%\todo{add why not benchmarking in ASL}

\section{\speechbelebele}
\label{sec:dataset}

\begin{table}[h!]
    \centering
    \scriptsize
    \begin{tabular}{ll|ll}
\toprule
 \multicolumn{2}{l}{Passages} & \multicolumn{2}{|l}{Questions/Answers} \\
\midrule

%\multicolumn{7}{l}{\belebele} \\
%\multicolumn{7}{l}{\speechbelebele} \\

 Distinct Passages & 488 & Distinct Q & 900 \\
 Questions per passage & 1-2 & Multiple-choice A & 4\\ 
 Avg words (std) &  79.1 (26.2) & Avg words Q (std) & 12.9 (4.0) \\
 Avg sentences (std) &  4.1 (1.4) & Avg words A (std) & 4.2 (2.9)\\ 
\bottomrule
    \end{tabular}
    \caption{Statistics for \speechbelebele, which covers \numlanguages spoken languages plus ASL. Average words are computed for English. \label{tab:statistics}}

\end{table}

\begin{figure}[ht!]
\center
    \includegraphics[width=7.5cm]{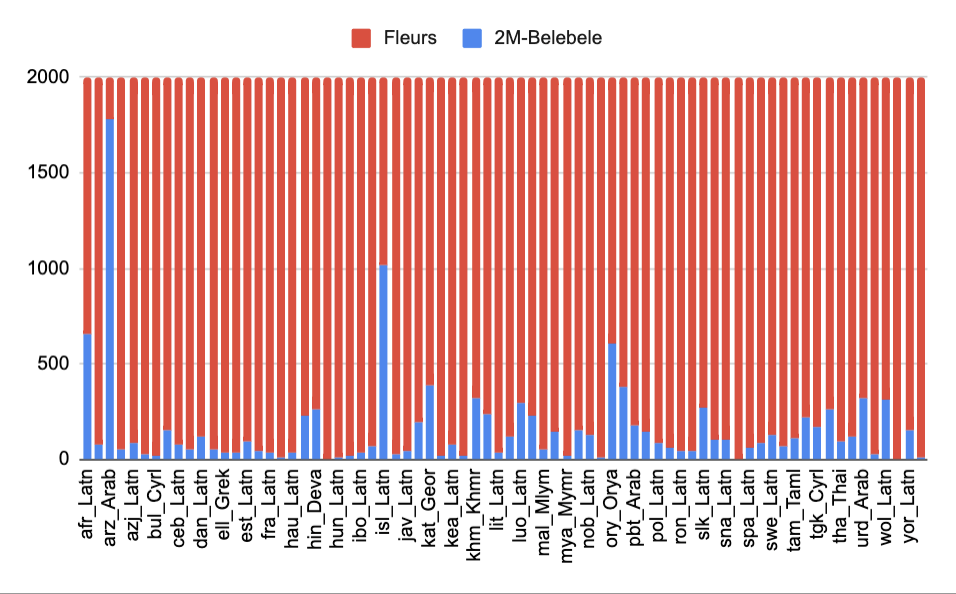}
    \caption{\fleurs vs New Recordings from \speechbelebele for sentences in passages. %Full \belebele questions and answers were recorded for covered languages.}
    \label{fig:statisticsfleursannotation}}
\end{figure}

\paragraph{\fleurs and \belebele passage alignment.} Since \belebele uses passages constructed from sentences in the \flores dataset, and \fleurs \citep{conneau2022fleurs} is a human speech version of \flores for a subset of its languages, we create a speech version of \belebele by aligning its passages with the speech segments available in \fleurs. This extension can be done without extra human annotation, just by computing the alignment between \fleurs and \belebele passages. However, such alignment does not cover the entire \belebele corpus because \fleurs does not cover the entirety of \flores. There are \numlanguages languages shared between \fleurs and \belebele. \fleurs does not cover the same passages as \belebele in all those 74 languages, which means that some languages have more speech passages than others. In general, we are able to match almost $\approx$ 80\% of the passages. Figure \ref{fig:statisticsfleursannotation} shows the number of \fleurs paragraphs we can match, thus obtaining the number of paragraphs that must be recorded in order to cover all passages \belebele. %Almost 50\% of the languages cover 90\% of the distinct passages and questions. Details of passage coverage and intersection across languages are reported in appendix \ref{sec:histogram}.

%\pierre{Further explain how alignment was done}

\paragraph{Speech recordings.}
We commission human recordings for the part of the \belebele dataset that is not covered by existing \fleurs recordings, as well as for elements of \belebele that do not exist in \fleurs (i.e. questions and answers). %Languages and speaker requirements for this task are \numlanguages languages and two speakers per segment. These speakers have to cover 1 higher-pitch (e.g., female) speaker and 1 lower-pitch (e..g., male) speaker for each passage.  
Recording participants must be native speakers of the  languages they record. They must have an impeccable grasp of the conventions used in their respective languages for the narration of texts. The three tasks that participants are asked to perform are: (1) Read aloud and record the text passages provided (from \flores); (2) Read aloud and record the provided written questions; (3) Read aloud and record the provided written answers.
For the task, we provide the participants with (a) the text of the sentences to be recorded in TSV format (the number of passages may differ from language to language), (b) the written questions (900 per language, and (c) the written answer options (3,600 per language). Additional details on the recording guidelines provided to annotators are reported in the appendix \ref{app:annot}.
We verify the quality of the recordings by randomly selecting 270 recordings (30\% of sample size) and ensuring that the recordings do not contain background or ambient noise and that the voices of the participants are clearly audible.

\paragraph{Sign recordings.}
To obtain ASL sign recordings, we provide translators of ASL and native signers with the English text version of the sentences to be recorded. The interpreters are then asked to translate these sentences into ASL, create glosses for all sentences, and record their interpretations into ASL one sentence at a time. The glosses are subjected to an additional quality check by expert annotators to harmonize the glossing format.  
To harmonize the recording conditions and eliminate visual bias, the videos are recorded against plain monochrome backgrounds (e.g., white or green), and signers are requested to wear monochrome upper body clothing (e.g., black). All videos are captured in 1920x1080p resolution with all of the signing space covered in FOV. The recordings are done in 60 frames per second to address most of the motion blur that happens during signing.

\paragraph{\speechbelebele Statistics.} The final dataset is composed of 75 languages (74 in speech, 1 in sign). Each of the languages' respective subsets includes 2,000 utterances organized in 488 distinct passages, 900 questions, and 4 multiple choice answers per question. For our recorded data (the red portion of Figure \ref{fig:statisticsfleursannotation} plus questions and answers), we have one audio file or two per sentence, depending on the number of available participants (one participant only in 23 languages, and two participants in 51 languages). When two speakers are available, we request that one should represent a higher-pitch range, and the other a lower-pitch range for each passage. More details are available in Appendix \ref{sec:appendix}.

In addition, the data set includes video recordings in ASL for 2,000 FLORES sentences (not including the test partition) and is similarly organized in 488 distinct passages, as well as 900 questions and 4 multiple-choice answers for each question (see summary table \ref{tab:statistics}). The ASL dataset was recorded by two interpreters, but, contrary to what was possible in other languages, each interpreter could only cover one-half of the dataset each. 

%\belen{do TTS for speech documents}

\section{Experiments}
\label{sec:experiments}

We evaluate \speechbelebele, and compare performance across modalities. Our comparison is limited in number of systems and combination of modalities. \speechbelebele offers the opportunity to check multimodal comprehension by combining speech/text/sign passages; questions and answers. In our case, we only provide results for entire text passages, questions and answers and speech passages, text questions and answers. A more comprehensive set of experiments is out of the scope of this paper, which aims at unblocking such experimentation by open-sourcing the dataset itself. 
%For the document-level experiments, we normalize both the passage and Wikipedia text (e.g., removing continuous spaces), then we search the passage in the Wikipedia text, if the passage can not be found in Wikipedia text, we insert the passage into a random position in the Wikipedia paragraphs. This occurs for 28 passages out of 488. %(54/900) Otherwise, we use the Wikipedia text as it is. Finally, we verify the length of the Wikipedia text.If the Wikipedia text is approximately longer than the maximal sequence length of systems on which we are benchmarking (e.g. \llamatwochat, 4096 tokens), we randomly drop the paragraphs either from the beginning or ending of the Wikipedia text while keeping the \flores passage. This occurs for 121 passages out of 488.% (225/900 with seed 0)}
%\bokai{any detail to add? }

\begin{figure}[ht!]
\center
    \includegraphics[width=7.5cm]{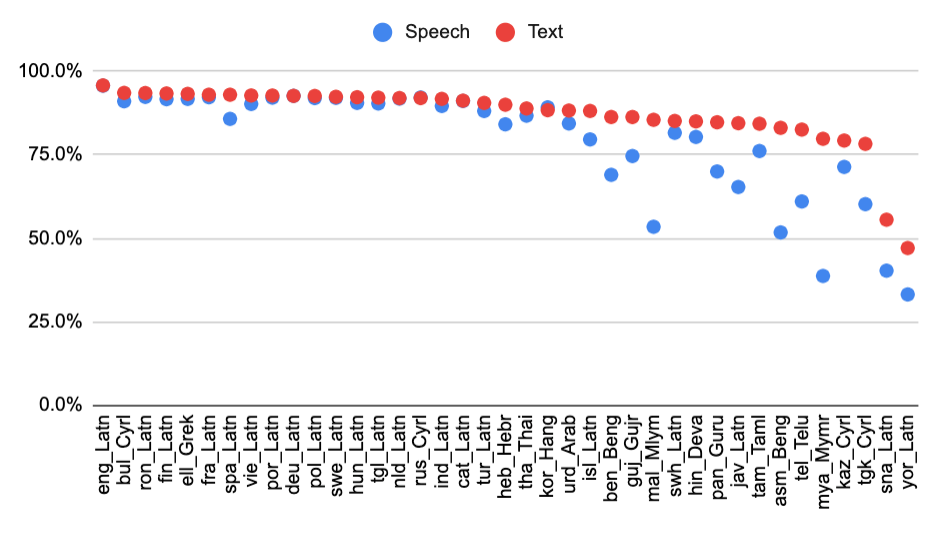}
    \caption{Speech and Text \belebele accuracy results in 39 languages. We compare text performance with %\llamathreebase (5-shot), \whispers+\llamathreebase (asr+5-shot), 
    \llamathreechat (zero-shot) and speech performance with \whispers+\llamathreechat (asr+zero-shot). %The latter includes question and answer in text and passage in speech. 
    \label{fig:langperf}}
\end{figure}

\begin{table*}[h!]
    \centering
    \scriptsize
    \begin{tabular}{ccc r r|c| c c | c c}
        \toprule
        \textbf{Dataset} &\textbf{Model} & \textbf{Size} & \textbf{Vocab} & \textbf{AVG} & \textbf{\% $\geq$ 50} & \textbf{\% $\geq$ 70} & \textbf{Eng} & \textbf{non-Eng} \\
        \midrule
        \multicolumn{9}{l}{\emph{5-Shot In-Context Learning (examples in English)}}  \\ \midrule
        %\belebele &\llamatwobase & 70B & 32K & 59.1 & 55.6 & 46.7 & 91.9 &  58.4\\  
         \belebele &\llamathreebase & 70B & 128K & 84.9 & 97.4 & 94.9 & 94.8 & 84.7 \\  
       
        %\speechbelebele&\whisper + \llamatwobase & 70B & 32K & 46.8 & 46.7 & 8.9& 77.1 & 46.1 \\
         \speechbelebele&\whispers + \llamathreebase & 70B & 128K & 77.1 & 89.7  &71.8& 94.4 & 76.6\\
         \speechbelebele&\mfourt + \llamathreebase & 70B & 128K & 81.7& 94.9 &92.7& 93.5 &  81.4\\
          \speechbelebele&\whispers + \llamatwobase & 7B & 32K & - & - &-& 49.9 & -\\
 %        \speechbelebele&\mfourt + \llamatwobase & 7B & 32K &  &  &&  &  \\
         %\speechbelebele &\whisper + \llamatwobase & 7B & 32K &  & 0 & 0 & 42.3 &     \\
         \speechbelebele&\spiritlm  & 7B & 37K & - & - &- & 25.9 & - \\ %\midrule 
%          \speechbelebele&\ssvpslt + \llamathreebase & 70B & 128K & - &  -&-& {\color{red}4/11} & - \\
        \midrule
        \multicolumn{9}{l}{\emph{Zero-Shot}}  \\ \midrule
        %\belebele &\llamatwochat & 70B & 32K & 47.1 & 42.2&  22.2&  81.8&  47.0\\  
         \belebele &\llamathreechat & 70B & 128K & 87.0& 97.4& 94.9 & 95.8 & 86.7 \\  
        %\speechbelebele &\whisper + \llamatwochat  & 70B & 32K & 39.1 & 33.3 & 2.2&  64.7& 38.5 \\ 
         \speechbelebele &\whispers + \llamathreechat  & 70B & 128K & 79.1& 92.3 & 76.9& 95.7 & 78.7 \\ 
          \speechbelebele &\mfourt + \llamathreechat  & 70B & 128K & 84.8 & 94.9 &94.9& 95.5 &  84.5\\  
%           \speechbelebele &\mfourt + \llamathreechat & 70B & 32K &  &  &&  &  \\ 
%\midrule 
 %           \speechbelebele & \ssvpslt + \llamathreechat & 70B & 128K & - & - &-&{\color{red}4/11}  & - \\
       % \multicolumn{9}{l}{\emph{Zero-Shot Document}}  \\
       % \belebele Document &\llamatwochat & 70B & 32K & -& -&  -&  82.2&  -\\ 
       % \belebele Document &\llamathreechat & 70B & 32K & -& -&  -&  &  -\\ 
          
       %  Speech \belebele Document &\whisper + \llamatwochat   & 70B & 32K & -& -&  -&77.4 &  -\\  
        % Speech \belebele Document &\whisper + \llamathreechat   & 70B & 32K & -& -&  -&&  -\\ 
        % Speech \belebele Document &\mfourt + \llamathreebase & 70B & 32K & - & - &-&  & - \\
        \bottomrule
    \end{tabular}
    \caption{Summary of accuracy results on \speechbelebele compared to \belebele across models and evaluation settings. \% $\geq$ 50/70 refers to the proportion
of languages for which a given model performs above 50/70\% for question and answer in text and passage in speech.      
    \label{table:summary}}    
\end{table*}

\paragraph{Systems.} We use the speech section of the  \speechbelebele dataset to evaluate the speech comprehension task with a cascaded system consisting of first speech recognition (ASR) using the \whisper model \citep{radford2022robust} (hereinafter, \whispers) and \mfourt (corresponding to \textsc{SeamlessM4T-Large v2}) model \cite{communication2023seamless} feeding into \llamathree\footnote{\url{https://ai.meta.com/blog/meta-llama-3/}}. We also provide results with a unified system \spiritlm \citep{nguyen2024spiritlm}, which is a multimodal language model that freely
mixes text and speech. Since the size of this model is 7B and is based on \llamatwo, we also add a comparison to the \llamatwobase model.  We compare these results with \llamathree and \llamathreechat using the \belebele text passage as input. For these systems, we report the results in 5-shot in-context learning and zero-shot on 39 languages at the intersection of \whispers, \mfourt and \speechbelebele (see Appendix \ref{sec:appendix}).  %We add experiments with best ASR and \llamatwobase to compare with \spiritlm, since this was built on top of this model. 

% \bokai{provide details on the systems and the evaluation modes, as in the original paper, see below}
% \paragraph{5-shot In-Context Learning.}
% %We evaluate \llamathreebase and \spiritlm in the 5-shot setting. 
% We use the same few-shot template as \cite{bandarkar2023belebele}, the few-shot examples are taken randomly from the English training set and they are prompted as \textit{text} format to the model. 

% \paragraph{Zero-shot Evaluation.}
% %We evaluate \llamathreechat in the zero-shot setting, 
% Again, we use the same evaluation strategy as in \cite{bandarkar2023belebele}. \spiritlm is not available in chat mode.%, therefore we only test on 5-shot strategy.
\paragraph{Zero-shot Evaluation.}
%We evaluate \llamathreechat in the zero-shot setting, 
We use the same evaluation strategy as in \cite{bandarkar2023belebele}. \spiritlm is not available in chat mode.
\paragraph{5-shot In-Context Learning.}
The few-shot examples are taken randomly from the English training set and they are prompted as \textit{text} format to the model. 
Different from \cite{bandarkar2023belebele}, we do not pick the answer with the
highest probability but directly assess the predicted letter of the answer.

For 5-shot and zero-shot settings, our instruction prompt is as follows \textit{“Given the following passage, query, and answer choices, output the letter corresponding to the correct answer. Do not write any explanation. Only output the letter within A, B, C, or D that corresponds to the correct answer.”} and we report the averaged accuracy over 3 runs\footnote{Random seeds: 0, 1, 2.}.

% {\color{red}
% \paragraph{5-shots In-Context Learning}

% We evaluate the pretrained  \llamatwobase in the 5-shots setting. Examples are sampled from the English training set and prompted to the model (following the template \texttt{P: <passage> \textbackslash{n} Q: <question> \textbackslash{n} A: <mc answer 1> \textbackslash{n} B: <mc answer 2> \textbackslash{n}  C: <mc answer 3> \textbackslash{n}  D: <mc answer 4> \textbackslash{n}  Answer: <Correct answer letter>}). We report the average scores over 3-runs. In this setting, we perform prediction by picking the answer within \{A, B, C, D\} that has the highest probability relatively to the others.

% \marta{only the passage is speech, the examples are text, the question and responses are text}

% \paragraph{Zero-shot Evaluation}

% We evaluate \llamatwochat in the zero-shot setting by describing the task in natural language. We present the passage, question, and four possible answers, and instruct the model to provide the letter ``A'', ``B'', ``C'' or ``D ''as the answer. The instructions are given in English for all languages.  We perform post-processing steps and accept answers predicted as e.g. ``(A)'' instead of ``A''.\footnote{For \llamatwochat we additionally remove the prefix ``The correct answer is ''.}

% }

\paragraph{Results.} \speechbelebele accuracy results per language compared to \belebele are shown in Figure \ref{fig:langperf}. Differences in speech and text vary slightly depending on the languages. Low-resource languages have a greater variation between text and speech \belebele. The ten languages with the largest gap are: Burmese, Maltese, Assamese, Telugu, Javanese, Tajik, Bengali, Shona, Eastern Panjabi, Yoruba, Gujarati. 
Table \ref{table:summary} reports the summary of the results. The English drop from direct text to speech task does not vary much between 5-shot and zero-shot strategies, being slightly higher in the zero-shot setting (coherently with previous \llamathree results that show better performance in zero-shot in other tasks\footnote{\url{https://ai.meta.com/blog/meta-llama-3-1/} and \url{https://ai.meta.com/blog/meta-llama-3/}}). %Note that highest drop between text and speech comprehension comes in the proportion of languages that perform above 70\%. 
When comparing speech and text comprehension, we observe that speech decreases performance in about 2-3\% average across languages depending on the setting.
Table \ref{table:summary} reports English results for \spiritlm, a direct multimodal model. One of the reasons \spiritlm may be performing worse is that 5-shot examples are in text, while the passage on the asked question is in speech. %Finally, Table \ref{table:summary} reports results on zero-shot (Speech) \belebele Document showing that the accuracy is slightly higher (+0.4) for text, but gains are larger (+12.7) for speech when adding larger context.

 %We used the self-supervised Video Pretraining for Sign Language Translation\footnote{https://github.com/facebookresearch/ssvp$\_$slt/}(\ssvpslt) \cite{rust-etal-2024-towards} to transcribe ASL. However, results were extremely poor. {\color{purple} Pierre, do we have some examples?} {\color{green} Cihan: We should add some numbers and some qualitative results if possible}. {\color{red} Belen verify the reference is the right one, please add any details} 

\paragraph{ASL} We know from previous large-scale translation attempts \citep{albanie2021bbc, muller2022findings} that models struggle to generalize over both individuals/appearance and large domain of discourse. Compared to speech and text models, sign language models suffer from having to learn generalized representations from high-dimensional inputs, i.e. videos, without overfitting to limited training dataset. Previous attempts have been made to create a more generalizable abstraction layer in the form of subunits \citep{camgoz2020multi}, similar to phonemes for speech, which achieved promising results on a translation task with a small discourse domain. However, this work is yet to be applied to large discourse domain translation tasks. The best results in the FLORES domain have been achieved with close models that are not available \citep{zhang2024scalingsignlanguagetranslation}. Trying \cite{rust-etal-2024-towards} as an open model did not perform above chance in the final reading comprehension dataset.
However, we believe that the release of this new dataset with the additional gloss annotation will help in training models that generalize over individuals better and improve large-scale sign language translation.

\section{Conclusions}
The \speechbelebele dataset\footnote{\speechbelebele dataset is freely available in Github \url{https://github.com/facebookresearch/belebele} and in HuggingFace \url{https://huggingface.co/datasets/facebook/2M-Belebele}} allows to evaluate natural language comprehension in a large number of languages, including ASL. \speechbelebele is purely human-made and covers the \belebele passages, questions, and answers for 75 languages: 74 in the speech modality and 1 in the sign modality. As a by-product, \mflores{} extends \fleurs by 20\% \footnote{\mflores is freely available in HuggingFace \url{https://huggingface.co/datasets/facebook/2M-Flores-ASL}}.%Additionally, we benchmark (Speech) \belebele Document extension for English.

\section*{Limitations and ethical considerations}

%Appendix \ref{sec:impact} reports an analysis of the impact of using TTS instead of human speeches.

%While Appendix \ref{sec:impact} reports the impact of using TTS instead of human speeches for building the speech comprehension evaluation dataset, it is only computed on English. Conclusions may vary in other languages. However, we still think that Speech \belebele v2 is valid to extract conclusions on speech comprehension, mainly because it counts on a large proportion (82\%) of human labeled speech. To solve this limitation, in the mid-term, we plan to release further versions of the extended \belebele with a higher proportion of human labeled speech. 
%Our document extension currently only covers English, in the future we plan to extend it to multiple languages either by aligning Wikipedia articles in other languages or by translating English documents.

Our speech annotations do not have the entire set completed with two annotators. Due to the high volume of the dataset, not every recording has been thoroughly verified. Some of the languages in \speechbelebele are low-resource languages, which pose a challenge in sourcing professionals to record. Therefore, some of the audios were recorded in home settings and may contain minor background noise, static noise, echoes, and, occasionally, the speech could be slightly muffled or soft. All annotators are native speakers of the target language, but they may have regional accents in their speech, and their personal speech styles may be present in the audio as well. However, the mentioned imperfections should not affect intelligibility; all the recordings can be clearly understood by human standards. Note that we are planning to release more languages as reported in Appendix \ref{app:morelang}.

We can group the ASL limitations under two categories, namely visual and linguistic. For visual limitations, ASL sequences are recorded in what can be considered laboratory environments with few signer variance. This makes it harder for models trained on them to generalize to unseen environments and signers. For linguistic limitations, ASL sequences are collected one sentence at a time. Although this enables pairwise training and evaluation, such as classical text-based NMT, the generated sequences may not be fully realistic in terms of real-world signing. An example would be the use of placement. In sentence-per-sentence sequence generation, a signer would refer to an entity with their sign each sentence, whereas in long-form conversation, a signer would place the entity in their signing space after first reference and refer them in via use of placement in the following sentences.

Our benchmarking is limited compared to the potential capabilities of the dataset. For example, since we have spoken questions, passages and responses, instead of just using a fix modality (spoken passages, text questions and responses), we could explore the performance when using all combinations among modalities (e.g., question in speech, answer in speech, passage in speech; or question in speech, answer in text, passage in speech; or question in speech, answer in speech and passage in text.)

In terms of compute budget, we estimate it as 47K Nvidia A100 hours by taking into account the product of following factors: number of languages (39), number of random seeds (3), number of GPUs required by model (8), number of experiment setups (5) and estimated number of hours per experiment (10).

Speakers and signers were paid a fair rate. Our recorded data reports self-identified gender by participant. Each of the speakers and signers signed a consent form agreeing on the dataset and its usage that they were participating in.

\section*{Acknowledgments}

This paper is part of the LCM project\footnote{https://github.com/facebookresearch/large$\_$concept$\_$models} and authors would like to thank the entire LCM team for the fruitful discussions.

%Authors want to thank Eduardo Sánchez for early discussions on the project.

%\bibliographystyle{abbrvnat}
\bibliography{anthology,custom}

\appendix

\section{Languages}
\label{sec:appendix}

Table \ref{table:languages} reports details on languages covered by \fleurs, TTS and ASR.
\begin{table*}
\small
\begin{tabular}{lllllcc}
    \toprule
    \textbf{Language} & \textbf{Code} & \textbf{Script}  & \textbf{Family} & \fleurs  & ASR &\speechbelebele \\ \midrule
    Mesopotamian Arabic&acm$\_$Arab&Arab&Afro-Asiatic& & \\  
Afrikaans&afr$\_$Latn&Latn&Indo-European&$\checkmark$& & $\checkmark$(1) \\ 
Tosk Albanian&als$\_$Latn&Latn&Indo-European& & \\ 
Amharic&amh$\_$Ethi&Ethi&Afro-Asiatic&$\checkmark$& & $\checkmark$(2) \\
North Levantine Arabic&apc$\_$Arab&Arab&Afro-Asiatic& & \\ 
Modern Standard Arabic&arb$\_$Arab&Arab&Afro-Asiatic& & \\ 
Modern Standard Arabic&arb$\_$Latn&Latn&Afro-Asiatic& & \\ 
Najdi Arabic&ars$\_$Arab&Arab&Afro-Asiatic& & \\ 
Moroccan Arabic&ary$\_$Arab&Arab&Afro-Asiatic& & \\
Egyptian Arabic&arz$\_$Arab&Arab&Afro-Asiatic&$\checkmark$ & &$\checkmark$(2) \\ 
Assamese&asm$\_$Beng&Beng&Indo-European&$\checkmark$&$\checkmark$& $\checkmark$(2) \\
North Azerbaijani&azj$\_$Latn&Latn&Turkic&$\checkmark$& & $\checkmark$(1) \\ 
Bambara&bam$\_$Latn&Latn&Niger-Congo& & \\
Bengali&ben$\_$Beng&Beng&Indo-European&$\checkmark$&$\checkmark$&$\checkmark$(2) \\
Bengali&ben$\_$Latn&Latn&Indo-European& & \\
Standard Tibetan&bod$\_$Tibt&Tibt&Sino-Tibetan& & \\
Bulgarian&bul$\_$Cyrl&Cyrl&Indo-European&$\checkmark$&$\checkmark$& $\checkmark$(2) \\
Catalan&cat$\_$Latn&Latn&Indo-European&$\checkmark$&$\checkmark$ & $\checkmark$(2) \\
Cebuano&ceb$\_$Latn&Latn&Austronesian&$\checkmark$ & & $\checkmark$(1) \\
Czech&ces$\_$Latn&Latn&Indo-European&$\checkmark$& & $\checkmark$(2) \\ 
Central Kurdish&ckb$\_$Arab&Arab&Indo-European&$\checkmark$& & \\  
Danish&dan$\_$Latn&Latn&Indo-European&$\checkmark$& & $\checkmark$(2) \\ 
German&deu$\_$Latn&Latn&Indo-European&$\checkmark$&$\checkmark$& $\checkmark$(2) \\
Greek&ell$\_$Grek&Grek&Indo-European&$\checkmark$&$\checkmark$& $\checkmark$(2) \\
English&eng$\_$Latn&Latn&Indo-European&$\checkmark$&$\checkmark$& $\checkmark$(2) \\
Estonian&est$\_$Latn&Latn&Uralic&$\checkmark$& & $\checkmark$(1) \\  
Basque&eus$\_$Latn&Latn&Basque& & \\
Finnish&fin$\_$Latn&Latn&Uralic&$\checkmark$&$\checkmark$& $\checkmark$(2) \\
French&fra$\_$Latn&Latn&Indo-European&$\checkmark$&$\checkmark$& $\checkmark$(2) \\
Fulfulde (Nigerian)&fuv$\_$Latn&Latn&Atlantic-Congo& & &\\ 
Oromo (West Central)&gaz$\_$Latn&Latn&Afro-Asiatic&($\checkmark$)& &\\ 
Guarani&grn$\_$Latn&Latn&Tupian& & \\
Gujarati&guj$\_$Gujr&Gujr&Indo-European&$\checkmark$&$\checkmark$& $\checkmark$(1) \\
Haitian Creole&hat$\_$Latn&Latn&Indo-European& & \\
Hausa&hau$\_$Latn&Latn&Afro-Asiatic&$\checkmark$& ($\checkmark$)& $\checkmark$(2) \\
Hebrew&heb$\_$Hebr&Hebr&Afro-Asiatic&$\checkmark$&$\checkmark$& $\checkmark$(2) \\
Hindi&hin$\_$Deva&Deva&Indo-European&$\checkmark$&$\checkmark$& $\checkmark$(2) \\
Hindi&hin$\_$Latn&Latn&Indo-European& & &\\ 
Croatian&hrv$\_$Latn&Latn&Indo-European&$\checkmark$& & $\checkmark$(2) \\
Hungarian&hun$\_$Latn&Latn&Uralic&$\checkmark$&$\checkmark$& $\checkmark$(2) \\
Armenian&hye$\_$Armn&Armn&Indo-European&$\checkmark$& & $\checkmark$(1) \\ 
Igbo&ibo$\_$Latn&Latn&Atlantic-Congo&$\checkmark$& & $\checkmark$(1) \\ 
Ilocano&ilo$\_$Latn&Latn&Austronesian& & \\
Indonesian&ind$\_$Latn&Latn&Austronesian&$\checkmark$&$\checkmark$& $\checkmark$(2) \\
Icelandic&isl$\_$Latn&Latn&Indo-European&$\checkmark$&$\checkmark$& $\checkmark$(1) \\
Italian&ita$\_$Latn&Latn&Indo-European&$\checkmark$& & $\checkmark$(2) \\ 
Javanese&jav$\_$Latn&Latn&Austronesian&$\checkmark$&$\checkmark$& $\checkmark$(1) \\
Japanese&jpn$\_$Jpan&Jpan&Japonic&$\checkmark$& & $\checkmark$(2) \\
Jingpho&kac$\_$Latn&Latn&Sino-Tibetan& & \\
Kannada&kan$\_$Knda&Knda&Dravidian&$\checkmark$& \\
Georgian&kat$\_$Geor&Geor&Kartvelian&$\checkmark$& & $\checkmark$(2) \\
Kazakh&kaz$\_$Cyrl&Cyrl&Turkic&$\checkmark$&$\checkmark$& $\checkmark$(1)\\
Kabuverdianu&kea$\_$Latn&Latn&Indo-European&$\checkmark$& & $\checkmark$(1) \\ 

 \bottomrule

    \end{tabular}
     \caption*{} 
\end{table*}

\begin{table*}
\small
\begin{tabular}{lllllcc}
    \toprule
    \textbf{Language} & \textbf{Code} & \textbf{Script}  & \textbf{Family} & \fleurs & ASR &\speechbelebele \\ \midrule
Mongolian&khk$\_$Cyrl&Cyrl&Mongolic&($\checkmark$)& & $\checkmark$(2) \\ 
Khmer&khm$\_$Khmr&Khmr&Austroasiatic&$\checkmark$& & $\checkmark$(1) \\
Kinyarwanda&kin$\_$Latn&Latn&Atlantic-Congo& & \\
Kyrgyz&kir$\_$Cyrl&Cyrl&Turkic&$\checkmark$& &\\
Korean&kor$\_$Hang&Hang&Koreanic&$\checkmark$&$\checkmark$& $\checkmark$(1) \\
Lao&lao$\_$Laoo&Laoo&Kra-Dai&$\checkmark$& \\
Lingala&lin$\_$Latn&Latn&Niger-Congo&$\checkmark$& &\\ 
Lithuanian&lit$\_$Latn&Latn&Indo-European&$\checkmark$& & $\checkmark$(2) \\
Ganda&lug$\_$Latn&Latn&Atlantic-Congo&$\checkmark$& & $\checkmark$(1) \\
Luo&luo$\_$Latn&Latn&Atlantic-Congo&$\checkmark$& & $\checkmark$(2) \\ 
Standard Latvian&lvs$\_$Latn&Latn&Indo-European&($\checkmark$)& & $\checkmark$(2) \\ 
Malayam&mal$\_$Mlym&Mlym&Dravidian&$\checkmark$&$\checkmark$& $\checkmark$(2) \\
Marathi&mar$\_$Deva&Deva&Indo-European&$\checkmark$& \\
Macedonian&mkd$\_$Cyrl&Cyrl&Indo-European&$\checkmark$& & $\checkmark$(2) \\ 
Maltese&mlt$\_$Latn&Latn&Afro-Asiatic&$\checkmark$& &\\
Maori&mri$\_$Latn&Latn&Austronesian&$\checkmark$& & \\ 
Burmese&mya$\_$Mymr&Mymr&Sino-Tibetan&$\checkmark$&$\checkmark$& $\checkmark$(2) \\
Dutch&nld$\_$Latn&Latn&Indo-European&$\checkmark$&$\checkmark$& $\checkmark$(2) \\
Norwegian Bokmål&nob$\_$Latn&Latn&Indo-European&$\checkmark$& & $\checkmark$(2) \\ 
Nepali&npi$\_$Deva&Deva&Indo-European&$\checkmark$& & $\checkmark$(2) \\ 
Nepali&npi$\_$Latn&Latn&Indo-European& & \\ 
Northern Sotho&nso$\_$Latn&Latn&Atlantic-Congo&$\checkmark$& &\\ 
Nyanja&nya$\_$Latn&Latn&Afro-Asiatic&$\checkmark$& \\
Odia&ory$\_$Orya&Orya&Indo-European&$\checkmark$& & $\checkmark$(1) \\
Eastern Panjabi&pan$\_$Guru&Guru&Indo-European&$\checkmark$&$\checkmark$& $\checkmark$(2) \\
Southern Pashto&pbt$\_$Arab&Arab&Indo-European&($\checkmark$)& & $\checkmark$(1) \\ 
Western Persian&pes$\_$Arab&Arab&Indo-European&($\checkmark$)& & $\checkmark$(1) \\ 
Plateau Malagasy&plt$\_$Latn&Latn&Austronesian& & \\ 
Polish&pol$\_$Latn&Latn&Indo-European&$\checkmark$&$\checkmark$& $\checkmark$(2) \\
Portuguese&por$\_$Latn&Latn&Indo-European&$\checkmark$&$\checkmark$& $\checkmark$(2) \\
Romanian&ron$\_$Latn&Latn&Indo-European&$\checkmark$&$\checkmark$& $\checkmark$(2) \\
Russian&rus$\_$Cyrl&Cyrl&Indo-European&$\checkmark$&$\checkmark$& $\checkmark$(2) \\
Shan&shn$\_$Mymr&Mymr&Tai-Kadai& & \\ 
Sinhala&sin$\_$Latn&Latn&Indo-European& & \\ 
Sinhala&sin$\_$Sinh&Sinh&Indo-European& & \\ 
Slovak&slk$\_$Latn&Latn&Indo-European&$\checkmark$& & $\checkmark$(1) \\ 
Slovenian&slv$\_$Latn&Latn&Indo-European&$\checkmark$& & $\checkmark$(2) \\ 
Shona&sna$\_$Latn&Latn&Atlantic-Congo&$\checkmark$&$\checkmark$& $\checkmark$(2) \\
Sindhi&snd$\_$Arab&Arab&Indo-European&$\checkmark$& & $\checkmark$(2) \\
Somali&som$\_$Latn&Latn&Afro-Asiatic&$\checkmark$\\
Southern Sotho&sot$\_$Latn&Latn&Atlantic-Congo& & \\ 
Spanish&spa$\_$Latn&Latn&Indo-European&$\checkmark$&$\checkmark$& $\checkmark$(2) \\
Serbian&srp$\_$Cyrl&Cyrl&Indo-European&$\checkmark$& & $\checkmark$(2) \\
Swati &ssw$\_$Latn&Latn&Atlantic-Congo& & \\ 
Sundanese &sun$\_$Latn&Latn&Austronesian& & \\
Swedish&swe$\_$Latn&Latn&Indo-European&$\checkmark$&$\checkmark$& $\checkmark$(2) \\
Swahili&swh$\_$Latn&Latn&Atlantic-Congo&$\checkmark$&$\checkmark$& $\checkmark$(1) \\
Tamil&tam$\_$Taml&Taml&Dravidian&$\checkmark$&$\checkmark$& $\checkmark$(2) \\
Telugu&tel$\_$Telu&Telu&Dravidian&$\checkmark$&$\checkmark$& $\checkmark$(2) \\
Tajik&tgk$\_$Cyrl&Cyrl&Indo-European&$\checkmark$&$\checkmark$& $\checkmark$(1) \\
Tagalog&tgl$\_$Latn&Latn&Austronesian&($\checkmark$)&$\checkmark$& $\checkmark$(2) \\
Thai& tha$\_$Thai &Thai&Tai-Kadai& $\checkmark$&$\checkmark$& $\checkmark$(2) \\
Tigrinya&tir$\_$Ethi&Ethi&Afro-Asiatic& & \\
Tswana&tsn$\_$Latn&Latn&Atlantic-Congo& & \\ 
 \bottomrule

    \end{tabular}
\end{table*}

\begin{table*}
\small
\begin{tabular}{lllllcc}
    \toprule
    \textbf{Language} & \textbf{Code} & \textbf{Script}  & \textbf{Family} & \fleurs & ASR &\speechbelebele \\ \midrule
    Tsonga&tso$\_$Latn&Latn&Afro-Asiatic& & \\
Turkish&tur$\_$Latn&Latn&Turkic&$\checkmark$&$\checkmark$& $\checkmark$(1) \\
Ukranian&ukr$\_$Cyrl&Cyrl&Indo-European&$\checkmark$& \\
Urdu&urd$\_$Arab&Arab&Indo-European&$\checkmark$&$\checkmark$& $\checkmark$(2) \\
Urdu&urd$\_$Latn&Latn&Indo-European& & \\ 
Northen Uzbek&uzn$\_$Latn&Latn&Turkic&$\checkmark$& & \\
Vietnamese&vie$\_$Latn&Latn&Austroasiatic&$\checkmark$&$\checkmark$& $\checkmark$(2) \\
Waray&war$\_$Latn&Latn&Austronesian& & \\
Wolof&wol$\_$Latn&Latn&Atlantic-Congo&$\checkmark$& & $\checkmark$(1) \\ 
Xhosa&xho$\_$Latn&Latn&Atlantic-Congo&$\checkmark$& & $\checkmark$(1) \\
Yoruba&yor$\_$Latn&Latn&Atlantic-Congo&$\checkmark$&$\checkmark$& $\checkmark$(2) \\
Chinese&zho$\_$Hans&Hans&Sino-Tibetan&$\checkmark$& & $\checkmark$(2) \\ 
Chinese&zho$\_$Hant&Hant&Sino-Tibetan&($\checkmark$)& \\ 
Standard Malay&zsm$\_$Latn&Latn&Austronesian&($\checkmark$)& & \\ 
Zulu&zul$\_$Latn&Latn&Atlantic-Congo&$\checkmark$& & \\ 
\midrule
American Sign Language&ase&-&Sign Language& & &$\checkmark$(2)\\
\bottomrule
   \end{tabular}
     \caption{Languages details. Column \fleurs reports the languages covered by Speech \belebele v1. Column ASR shows the languages reported in the experiment section, note that Hausa is covered by \whisper but not for \mfourt. The number in brackets shows the number of annotations per language. \label{table:languages} } 
\end{table*}

\section{Annotation Guidelines}
\label{app:annot}

\paragraph{Recording process.}
Find a quiet place free from distractions and noises, and choose a headphone that is comfortable to wear and a good quality microphone that will not distort or break your voice.
Read aloud and record the scripts in a pleasant tone and at a constant and even pace, as if you were reading a formal document. Try not to speak too quickly or slowly and aim for a natural pace that is easy to follow. The audio files below provide examples of paces that are expected, too fast, or too slow, for the sentence. The hearing also marks the date for the suspect’s right to a rapid trial.

To achieve the best sound quality when recording, position the microphone close to your mouth so that the voice will sound clear and present, but not too close that it sounds muddy or you can hear a puff of air. 
Clearly enunciate the words and avoid mumbling. Be sure to provide a 2-second pause between sentences to add clarity and keep the overall pace down.
When dealing with long, complicated sentences that contain multiple clauses or phrases, there are several approaches to ensure clarity and a natural flow as follows.
Break it down: Separate the sentence into smaller parts or clauses.
Practice reading aloud several times before starting the recording. This can help you get a feel for the rhythm and pacing of the sentence.
Pace yourself: Try to maintain a steady, even pace. If the sentence is particularly long, it is possible to take a brief pause at a natural breakpoint to catch your breath.
You should read the provided passages aloud without repairs (a repair is the repetition of a word that was incorrectly pronounced to correct its pronunciation).

To achieve this, familiarize yourself beforehand with the correct pronunciation of difficult words, proper nouns, and transliterated words, as well as signs and symbols, dates and times, numbers, abbreviations, and punctuation marks. Some elements may have more than one correct pronunciation. In this case, use the one that comes the more naturally to you, as long as it is an accepted pronunciation (i.e., it is acknowledged in your language's dictionaries). Practice reading the passages aloud several times to become more comfortable with the material. Please pay particular attention to the following items:

\paragraph{Numbers.} Number formats can vary from language to language; it is important to follow the pronunciation rules in your language. Here are some general guidelines and examples:
Decimal numbers: Read the whole part of the number as a whole number and then individually read every number after the decimal point. For example, in English, the decimal number 3.14 should be read as "three point one four." Different languages may have different rules, and you should follow the rules that are appropriate for your language.
Cardinal numbers represent quantities or amounts. Ordinal numbers represent positions or ranks in sequential order and should be read with the appropriate suffix. For example, in English, the ordinal number 1st is read "first" (not "onest") and 5th is read "fifth" (not "fiveth"). Different languages may have different rules, and you should follow the rule that is appropriate for your language.

Roman numerals are a collection of seven symbols that each represent a value: I = 1, V = 5, X = 10, L = 50, C = 100, D = 500, and M = 1,000. The can be pronounced in slightly different ways depending on the context, but they are never pronounced as individual letters. For example, in English, VIII in Henry VIII is pronounced "Henry the eighth", while Superbowl LVIII is pronounced "Superbowl fifty-eight", but they are never pronounced "Henry v i i i" or "Superbowl l v i i i". Different languages may have different rules, and you should follow the rules that are appropriate for your language.
Punctuation marks: As a general rule, punctuation marks should not be pronounced, except quotation marks.

For example, in English, punctuation marks such as periods, commas, colons, semicolons, question marks, and exclamation points are typically not pronounced. For example, the sentence. As a result of this, a big scandal arose. will be pronounced "As a result of this a big scandal arose" - not "As a result of this comma a big scandal arose period".
However, in formal-register English (in the news, for example), a difference is made between content created by the news team and content that should be attributed to someone else by explicitly pronouncing quotation marks. For example, the news transcript The fighter said: "I am here to try to win this." will be pronounced: "The fighter said, quote, I am here to try to win this. End of quote." In this case, different languages may have different rules, and you should follow the rules that are appropriate for your language.
Signs and symbols. Signs and symbols need to be pronounced as they would be heard in a speech-only setting. Attention should be paid:
(a) to potential number or gender agreement (for example, in English, "40\%" should be read as "forty percent" — not "forty percents") 
(b) to potential differences between the place of the sign or symbol in writing and in speech (for example, in English, the "\$" sign should be read as "dollar" and should be read after the number it precedes; i.e. "\$22" should be read as "twenty-two dollars" — not "dollars twenty-two") 
(c) to the way the sign or symbol gets expanded in speech (for example, in English, "Platform 9 ¾" should be read "platform nine and three quarters" — not "platform nine three quarters"). Similarly, 50 km/h would be pronounced "fifty kilometers per hour" — not "fifty kilometers hour").
Different languages may have different rules, and you should follow the rules that are appropriate for your language.

\paragraph{Proper nouns and foreign expressions.} Even the same language may have at least 2 different ways to pronounce foreign expressions of proper nouns:
(a) one way is to try to approach the way they would sound in the foreign language from which they come (for example, in English, Louis in Louis XIV is pronounced "lewee" as it would be in French);
(b) the other way is to pronounce them according to the rules of the adopting language (for example, in English, Louis in the City of St Louis is pronounced as in the English proper noun "Lewis")

\paragraph{Abbreviations.}  Abbreviations should be expanded as much as possible. However, it is suggested to refrain from expanding them if their expansion results in unnatural speech. For example, in English, abbreviations such as Dr. or etc. are pronounced "doctor" and "et cetera", respectively (not "d r" nor "e t c").  However, abbreviations such as AM or PhD are pronounced as a sequence of letters without being expanded ("a m" and "p h d", respectively - not "ante meridiem" nor "philosophy doctorate"). Different languages may have different conventions, and you should follow the conventions that are appropriate for your language.

\section{Extra languages pending for collection}
\label{app:morelang}

We plan to collect in total 91 languages with both high-pitched and low-pitched. This is the list of all the languages in planning.
\begin{itemize}
    \item Central Kurdish
    \item Nigerian Fulfulde
    \item West Central Oromo
    \item Kannada
    \item Kyrgyz
    \item Lao
    \item Lingala
    \item Marathi
    \item Maltese
    \item Maori
    \item Northern Sotho
    \item Chewa
    \item Somali
    \item Ukrainian
    \item Northern Uzbek
    \item Malay
    \item Zulu
\end{itemize}

%\appendix

%\section{\belebele Document Extension}

\section{Ablation study: Synthetic extension in speech evaluation datasets}
\label{sec:ablation}

In this part of our work, we aim to analyze the feasibility of synthetically extending text benchmarks to speech using TTS systems, thereby creating multi-modal datasets. Our goal is to understand if it would have been feasible to obtain the speech version of \belebele by using state of the art TTS systems, instead of human recordings. 

For this study we use \fleurs dataset, that contains ASR data in the same domain as \belebele. We chose to perform this study in the ASR task because it is simpler compared to other speech tasks, due to its monotonic alignment process and minimal need for reasoning. This ensures that the overall model performance and the complexity of the task are less likely to influence the results.

For our experiments, we generate a synthetic copy of the \fleurs dataset using the \mms TTS \citep{pratap2024scaling} system on the \fleurs transcripts. Then, we benchmark state-of-the-art models (\whispers, \mfourt and \mms ASR) on both the original and synthetic datasets and analyze whether the conclusions remain consistent across both datasets. \footnote{Note that we perform the study on the \fleurs languages that are covered by all \mms, \whispers and \mfourt.} 

It is important to note that a decrease in system performance is expected when using synthetic data. However, if this decrease occurs proportionally across all models, the synthetic data could still be useful to benchmark models. Conversely, if the model performance ranking changes, we can conclude that synthetic data is not reliable when benchmarking models.

To measure the variability in model rankings between the original and the synthetic data, we track the inversions that occur in the order of the models in the two settings. We define an inversion as a swap between two models that appear in adjacent positions on the list. We count how many swaps are needed in the ranking obtained using synthetic data to match the ranking from the original dataset.

% \begin{figure*}[t]
% \centering
% \includegraphics[width=0.8\textwidth]{figures/speech-eval.png}
% \caption{Number of Inversions vs Language Resources.}
% \label{fig:pipeline}
% \end{figure*}

\begin{table}[ht]
\scriptsize
\centering
\begin{tabular}{l|cc|cc|cc|c}
\hline
 & \multicolumn{2}{|c|}{\textbf{\mfourt}} & \multicolumn{2}{|c}{\textbf{\whispers}} & \multicolumn{2}{|c}{\textbf{\mms}} \\ %\cline{2-7}
 & \textbf{Hum} & \textbf{Syn} & \textbf{Hum} & \textbf{Syn} & \textbf{Hum} & \textbf{Syn} & \textbf{Inv}\\ \hline
Bengali & 14.1 & 21.1 & 114.7 & 105.8 & 14.6 & 25.0 &  \\
Catalan & 8.2 & 13.2 & 6.7 & 16.4 & 10.3 & 21.8 & \checkmark \\
Dutch & 9.9 & 20.0 & 8.5 & 19.7 & 12.4 & 28.3 &  \\
English & 6.0 & 11.7 & 4.5 & 9.8 & 12.3 & 19.2 &  \\
Finnish & 20.1 & 20.8 & 12.5 & 18.9 & 13.1 & 18.4 & \checkmark \\
French & 9.5 & 10.8 & 6.7 & 11.3 & 12.4 & 16.6 & \checkmark \\
German & 8.5 & 13.9 & 5.2 & 12.3 & 10.5 & 20.8 &  \\
Hindi & 11.9 & 13.4 & 33.5 & 28.7 & 11.1 & 18.3 & \checkmark \\
Indonesian & 12.1 & 12.8 & 8.7 & 14.2 & 13.2 & 21.9 & \checkmark \\
Korean & 25.7 & 40.3 & 15.4 & 29.9 & 47.8 & 61.2 &  \\
Polish & 13.0 & 14.7 & 8.1 & 13.3 & 11.6 & 18.1 & \checkmark \\
Portuguese & 9.0 & 8.0 & 4.1 & 6.9 & 8.7 & 10.4 & \checkmark \\
Romanian & 12.6 & 11.7 & 13.5 & 25.4 & 12.0 & 15.4 & \checkmark \\
Russian & 10.2 & 18.6 & 5.6 & 17.4 & 18.8 & 34.3 &  \\
Spanish & 6.3 & 9.1 & 3.4 & 10.0 & 6.4 & 10.8 & \checkmark \\
Swahili & 19.5 & 19.0 & 64.2 & 58.4 & 14.2 & 19.0 & \checkmark \\
Swedish & 15.4 & 20.1 & 11.3 & 19.1 & 21.0 & 27.8 &  \\
Telugu & 27.4 & 28.0 & 132.2 & 133.9 & 24.2 & 27.8 &  \\
Thai & 127.8 & 135.5 & 104.0 & 121.3 & 99.8 & 99.9 &  \\
Turkish & 18.6 & 23.0 & 8.4 & 16.5 & 19.2 & 30.3 &  \\
Ukrainian & 15.0 & 23.5 & 9.8 & 21.8 & 18.1 & 34.7 &  \\
Vietnamese & 16.0 & 20.1 & 10.2 & 14.2 & 25.8 & 25.3 &  \\ \bottomrule
\end{tabular}
\caption{WER($\downarrow$) results on the ASR task. Last column marks if the language has at least 1 inversion in ASR performance ranking comparing human vs TTS inputs.}
\label{tab:speech-results}
\end{table}

In Table \ref{tab:speech-results} we see that in the ASR setting, conclusions regarding model performance can vary depending on whether human or synthetic data is used. Although these conclusions are specific to the evaluated tasks and datasets, we demonstrate that even with the outstanding performance of current TTS methods, this does not guarantee the reliability of the data they generate when it comes to evaluation purposes. This is true not only for low-resource languages, but also for high-resource languages such as French or Spanish. These findings show that speech benchmarks might not be reliable if synthetically generated even in widely researched areas, further supporting the creation of evaluation datasets by humans.

\end{document}